\documentclass[11pt,a4paper]{article}

\usepackage[utf8]{inputenc}
\usepackage{multicol}

\usepackage{subfigure}
\usepackage[round,authoryear]{natbib}

\usepackage{bm}
\usepackage{amsmath}

\usepackage{boxedminipage}

\usepackage{listings}
\usepackage{color}
\definecolor{codegreen}{rgb}{0,0.6,0}
\definecolor{codegray}{rgb}{0.5,0.5,0.5}
\definecolor{codepurple}{rgb}{0.58,0,0.82}
\definecolor{backcolour}{rgb}{0.95,0.95,0.92}
\definecolor{codered}{rgb}{0.5,0,0}

\lstdefinestyle{mystyle}{
    backgroundcolor=\color{backcolour},   
    commentstyle=\color{codered},
    keywordstyle=\color{magenta},
    numberstyle=\tiny\color{codegray},
    stringstyle=\color{codegreen},
    breakatwhitespace=false,         
    breaklines=true,                 
    captionpos=b,                    
    keepspaces=true,                 
    numbers=left,                    
    numbersep=5pt,                  
    showspaces=false,                
    showstringspaces=false,
    showtabs=false,                  
    tabsize=2,
    basicstyle=\ttfamily\footnotesize
}
 
\usepackage{ifpdf}

\usepackage{tgpagella}
\usepackage[T1]{fontenc}
\usepackage{fancyhdr}   
\usepackage{makeidx}

\usepackage{lipsum}

\ifpdf
\usepackage[pdftex]{graphicx}
\else
\usepackage{graphicx}
\fi
\usepackage{rotating}
\usepackage{blindtext}
\usepackage[includemp=false, marginparwidth=1.8cm, hmargin=3.5cm, vmargin=3cm]{geometry}
\ifnum\pdfoutput>0			
\usepackage[pdftex, 			
pagebackref = false, 			
bookmarks = true, 			
bookmarksnumbered = false, 	
hyperindex = true,			
linktocpage = true,			
pdfpagemode = UseNone, 		
pdfstartview = Fit, 			
pdfpagelayout = OneColumn,
colorlinks = true, 			
urlcolor = blue, 				
citecolor = blue,
pdfborder = {0 0 0} 			
]{hyperref} 					
\fi

\ifpdf
\DeclareGraphicsExtensions{.png, .pdf, .jpg, .tif}
\else
\DeclareGraphicsExtensions{.eps, .jpg}
\fi

\usepackage{amsmath, amsthm}
\usepackage{amsfonts}
\usepackage{amssymb}

\title{{\bf Psychlab: A Psychology Laboratory for Deep Reinforcement Learning Agents}}

\author{\parbox{\linewidth}{\centering Joel Z. Leibo, Cyprien de Masson d'Autume, Daniel Zoran, David Amos, Charles Beattie, Keith Anderson, Antonio García Castañeda, \newline Manuel Sanchez, Simon Green, Audrunas Gruslys, Shane Legg, \newline Demis Hassabis, and Matthew M. Botvinick}}
\date{DeepMind, London, UK \\ \today}

\begin{document}

\maketitle

\begin{abstract}
{\bf Psychlab is a simulated psychology laboratory inside the first-person 3D game world of DeepMind Lab~\citep{beattie2016deepmind}. Psychlab enables implementations of classical laboratory psychological experiments so that they work with both human and artificial agents. Psychlab has a simple and flexible API that enables users to easily create their own tasks. As examples, we are releasing Psychlab implementations of several classical experimental paradigms including visual search, change detection, random dot motion discrimination, and multiple object tracking. We also contribute a study of the visual psychophysics of a specific state-of-the-art deep reinforcement learning agent: UNREAL~\citep{jaderberg2016reinforcement}. This study leads to the surprising conclusion that UNREAL learns more quickly about larger target stimuli than it does about smaller stimuli. In turn, this insight motivates a specific improvement in the form of a simple model of foveal vision that turns out to significantly boost UNREAL's performance, both on Psychlab tasks, and on standard DeepMind Lab tasks. By open-sourcing Psychlab we hope to facilitate a range of future such studies that simultaneously advance deep reinforcement learning and improve its links with cognitive science.}
\end{abstract}


\tableofcontents

\newcommand{\Ssp}{\ensuremath{\mathcal{S}}}
\newcommand{\Asp}{\ensuremath{\mathcal{A}}}
\newcommand{\Hist}{\ensuremath{\mathcal{H}}}



\section{Introduction}

State-of-the-art deep reinforcement learning (RL) agents can navigate 3D virtual worlds viewed from their own egocentric perspective~\citep{mirowski2016learning, wang2016learning, duan2016rl, chaplottransfer2016}, bind and use information in short-term memory~\citep{chiappa2017recurrent, vezhnevets2017feudal}, play "laser tag"~\citep{jaderberg2016reinforcement, chaplot2017arnold}, forage in naturalistic outdoor environments with trees, shrubbery, and undulating hills and valleys~\citep{purves2016}. Deep RL agents have even been demonstrated to learn to respond correctly to natural language commands like "go find the blue balloon"~\citep{hermann2017grounded, hill2017understanding, chaplot2017gated}. If they can do all that, then it stands to reason that they could also cope with the experimental protocols developed in the fields of psychophysics and cognitive psychology. If so, a new point of contact could be established between psychology and modern AI research. Psychology stands to gain a new mode for empirically validating theories of necessary aspects of cognition, while AI research would gain a wealth of tasks, with well-understood controls and analysis methods, thought to isolate core aspects of cognition.

In this work we contribute a new research platform: \emph{Psychlab}. This framework allows deep RL agents and humans to be directly compared to one another on tasks lifted directly from cognitive psychology and visual psychophysics. Psychlab was built on top of DeepMind Lab (henceforth "DM-Lab")~\citep{beattie2016deepmind}, the agent testing environment where much of the state-of-the-art in deep RL has been developed. This means that state-of-the-art agents can be directly plugged into Psychlab with no change to their setup. 

We also contribute Psychlab implementations of several classical experimental paradigms including visual search, multiple object tracking, and random dot motion discrimination. Human results on each of these paradigms look similar to standard results reported in the psychology literature, thus validating these implementations. 

Psychlab can be used for research that compares deep RL agents to human results in such a way that enriches agent understanding and thereby contributes back to improve agent design. As an example for how such a research program may unfold, in the second half of this paper we describe a set of experiments probing visual psychophysics of a deep RL agent. One result of these experiments was to motivate a particular agent improvement: a simple model of foveal vision that then improves performance on a range of standard DM-Lab tasks.

\begin{figure}[h!]
\includegraphics[width=\textwidth]{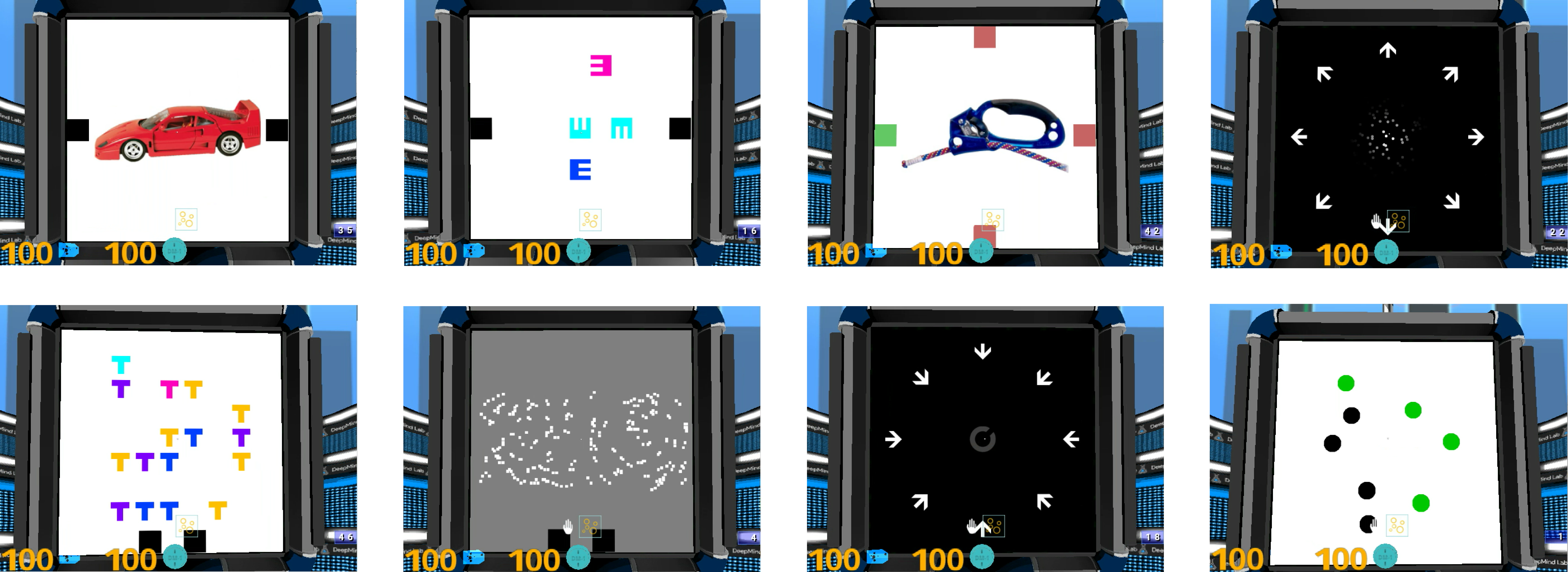}
\caption{Screenshots from the various tasks we are releasing with the Psychlab framework. Proceeding clockwise starting from the upper left, they are (1) continuous recognition, (2) change detection (3) arbitrary visuomotor mapping, (4) random dot motion discrimination, (5) visual search, (6) glass pattern detection, (7) Landolt C identification, and (8) multiple object tracking.}
\label{fig:gallery}
\end{figure}

\section{The Psychlab platform}

\subsection{Motivation}

Recently, AI research has been driven forward by the availability of complex 3D naturalistic environments like DeepMind Lab. These environments have been important for pushing artificial agents to learn increasingly complex behaviors. However, the very properties that make them attractive from the perspective of agent training also make it harder to assess what agents have in fact ended up learning. They do not unambiguously tell us what cognitive abilities agents that do well on them have developed. Moreover, while human benchmarking experiments can be performed in such environments, it is difficult even with humans to pin down the particular cognitive abilities involved in their successful performance. This is because the tasks generally depend on multiple abilities and/or admit multiple solution strategies. As a result, analyzing and understanding of artificial agents in relation to concepts from cognitive science has been challenging. 

Such ambiguities in the analysis of natural behavior have long been known in fields like psychology where the subject of interest is human. The answer developed over the last 150 years in psychology has been to design rigorously controlled laboratory-based experiments aimed at isolating one specific cognitive faculty at a time. Deep RL has now advanced to a point where it is productive to apply this same research methodology to tease apart agent cognitive abilities. Of course, it would be a huge undertaking to try come up with well-targeted and rigorously controlled experiments for all the cognitive functions we would like our agents to learn. Fortunately, we need not reinvent the wheel. If we set up our testing environment in the right way, i.e., like a typical psychology laboratory, then we can use exactly the same tasks that cognitive psychology has already invented. These tasks have already been validated by large bodies of previous research in that field.

On the other hand, there is a long history of modeling cognitive faculties or other psychological phenomena with neural networks (e.g.~\cite{mcculloch1943logical, rumelhart1987parallel}). However, these models typically do not work on raw sensory data or interact with laboratory equipment like the computer monitor and mouse as humans do. Thus they cannot actually be tested with the same experimental procedures that led to the conceptual results on which they are based\footnote{For example, the HMAX model of object recognition and the ventral visual pathway~\citep{riesenhuber1999hierarchical} is a feedforward neural network, and thus is conventionally interpreted as a model of object recognition performance under brief presentation conditions. Psychophysical experiments relating to it generally present stimuli for less than 100ms and employ backwards masking techniques thought to minimize the effects of feedback on visual processing~\citep{serre2007feedforward}. But the model itself just takes in images and outputs class labels. It has no notion of time.}. Moreover, these methods cannot actually be applied to the most advanced agents and algorithms developed through the complex and naturalistic approach to deep RL (e.g.~\cite{jaderberg2016reinforcement}).

Psychlab enables direct comparisons between agent and human cognitive abilities by making sure the protocols used to measure and assess performance are consistent for both. Moreover, since Psychlab tasks can be versions of classical behavioral paradigms that have stood the test of time, Psychlab has potential to offer better experimental controls and greater focus on probing specific cognitive or perceptual faculties.

\subsection{The Psychlab environment}\label{section:psychlab}

Psychlab is a psychophysics testing room embedded inside the 3D game world of DM-Lab. The agent stands on a platform in front of a large "computer monitor" on which stimuli are displayed. The agent is able to look around as in a usual 3D environment. It can decide to look at (fixate on) the monitor or around it e.g. down at the ground or up at the sky. Any change in gaze direction transforms the visual scene as projected in the agent's viewpoint in the usual way of a 3D game environment. That is, changes in gaze direction produce a global transformation of the visual scene. 

As is common in experiments with nonhuman primates and eye-tracking, the agent responds by "saccading" to (looking at) target stimuli. When humans use Psychlab they control their avatar's direction of gaze with the computer mouse. This is the exact same way human control works in many popular videogames such as first-person shooter games like Quake3, the game from which DM-Lab evolved. At least for subjects with experience playing such videogames, these controls feel quite natural and intuitive.

\begin{figure}[h]
\includegraphics[width=\textwidth]{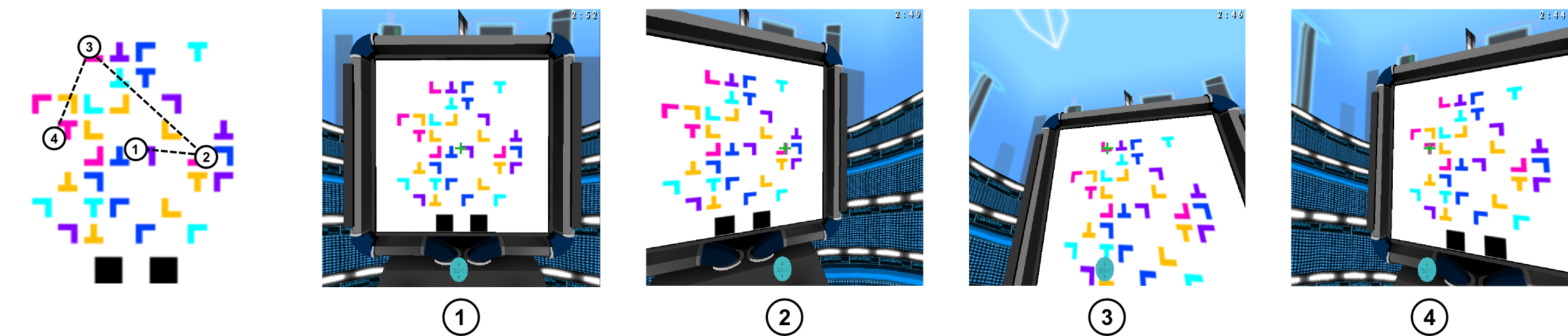}
\caption{A visualization of the visual search task as it looks in Psychlab. In the first snapshot (1) the agent is fixating at the center. In subsequent snapshots the target is searched for (2 and 3) and finally found, a magenta 'T', in (4).}
\label{fig:visualizaion}
\end{figure}

Psychlab comes with a flexible and easy-to-use GUI scripting language to control how stimuli are placed on the simulated computer monitor and program tasks. For readers who are familiar with software for coding behavioral tasks, Psychlab can be thought of as analogous to a scripting environment like  Psychtoolbox~\citep{brainard1997psychophysics}, but specialized for comparisons between deep RL agents and humans.

The reader is encouraged to take a look at any of the attached videos for examples of what human behavior in Psychlab looks like.

\subsection{Example Psychlab tasks}\label{section:example_tasks}

In keeping with common behavioral testing methods, all the Psychlab tasks we implemented were divided into discrete \emph{trials}. The trial was also the basic unit of analysis. There could be any number of trials in a DM-Lab \emph{episode} and both numbers of trials and episode duration can be configured by the experimenter. In all example tasks, a trial is initiated by fixating a red cross in the center of the Psychlab screen.

\subsubsection*{List of example Psychlab tasks and videos showing human performance}
\begin{enumerate}
    \item Continuous recognition \url{https://youtu.be/Yd3JhXC0hIQ}, probe recognition with a growing set of 'old' items.
    \item Arbitrary visuomotor mapping \url{https://youtu.be/71FvjgZXbF8}, cued recall task testing  memory for a growing list of item-response pairings.
    \item Change detection (sequential comparison) \url{https://youtu.be/JApfKhlrnxk}, subject must indicate if an array of objects that reappears after a delay has been changed.
    \item Visual acuity and contrast sensitivity (Landolt C identification) \url{https://youtu.be/yWA1-hFU9F0}, identify the orientation of a Landolt C stimulus appearing at a range of scales and contrasts.
    \item Glass pattern detection \url{https://youtu.be/XLQ9qgV9qrE}, subject must indicate which of two patterns is a concentric Glass pattern.
    \item Visual search \url{https://youtu.be/Vss535u4A5s}, subject must search an array of items for a target.
    \item Random dot motion discrimination \url{https://youtu.be/IZtDkryWedY}, subject must indicate the direction of coherent motion.
    \item Multiple object tracking \url{https://youtu.be/w3ddURoeQNU}, tests the ability to track moving objects over time.
\end{enumerate}

The default stimulus set for the continuous recognition and arbitrary visuomotor mapping tasks is from \cite{brady2008visual}.

\begin{figure}[h]
\begin{lstlisting}[language={[5.2]Lua}]
-- Create a table to contain Psychlab environment data
local env = {}

-- Function to initialize the environment, called at episode start
function env:_init(api, options)
    -- Create a widget that rewards the agent for looking at it
    local function lookCallback(name, mousePos, hoverTime, userData)
      api:addReward(REWARD_VALUE)
    end
    api:addWidget{
        name = WIDGET_NAME,
        image = WIDGET_IMAGE,
        pos = WIDGET_POSITION,
        size = WIDGET_SIZE,
        mouseHoverCallback = lookCallback,
    }

    -- Create a timer that removes the widget after some time
    local function timeoutCallback()
      api:removeWidget(WIDGET_NAME)
    end
    api:addTimer{
        name = TIMER_NAME,
        timeout = TIME_TILL_TIMEOUT,
        callback = timeoutCallback
    }
end

-- Construct DM-Lab level API around the Psychlab environment
local psychlab_factory = require "factories.psychlab.factory"
return psychlab_factory.createLevelApi{
    env = require "frameworks.point_and_click",
    envOpts = {environment = env}
}
\end{lstlisting}
\caption {Lua API example. This example places a widget on the screen. The subject is rewarded for looking at this widget. It disappears after some period of time.
\label{fig:lua_example} }
\end{figure}

\subsection{The Psychlab API}\label{section:api}

The Psychlab API is a simple GUI framework scripted in lua. Tasks are created by placing widgets on the Psychlab monitor. Widgets can have arbitrary visuals, and can invoke callbacks when events occur, such as the agent's center of gaze entering or exiting the widget area. The framework also supports timers that invoke callbacks when they complete.

\subsection{Reinforcement learning}\label{section:reinforcement_methods}

Psychlab supports and reward scheme the task developer can code using the API. In our experiments we provided rewards as follows: Agents receive a reward of 1 whenever they correctly complete a trial and a reward of 0 on all other steps. In our initial experiments we tried several other reward schemes, such as providing negative rewards for incorrect answers and small positive rewards for reaching basic trial events like foveating the fixation cross. However, we found that these led to slower and less stable learning.

\subsection{Relating protocols for humans, animals, and deep RL agents}

In Psychlab, reaction times are measured in terms of game steps. When humans use Psychlab, the engine runs at 60 steps per second. If the engine were run instead at 30 steps per second then human response times would be different. Thus it is only appropriate to make qualitative comparisons between human and agent response times with Psychlab.

Psychlab experimental protocols with deep RL agents can be seen as most directly comparable to non-human primate experimental protocols. This is because human psychophysics typically relies on conveying the task to the subject verbally. Since the deep RL agents we aim to study with Psychlab do not understand language, the tasks must be conveyed by reinforcement learning. Non-human primate psychophysics protocols are also conveyed by reinforcement learning for the same reason. 

The analogy between Psychlab protocols for testing deep RL agents and non-human primate protocols may run deeper still. One interpretational caveat affecting many non-human primate experiments is that, relative to the human versions of the same protocols, the monkey subjects may have been "overtrained". Since non-human primate training procedures can take many months, it's possible that slower learning mechanisms might influence results with non-human primates that could not operate within the much shorter timeframe (typically one hour) of the equivalent human experiment. In particular, perceptual learning~\citep{fine2002comparing} and habitization~\citep{seger2011critical} processes are known to unfold over such longer timescales. Thus they may be expected to play a role in the non-human primate protocols (see e.g.~\cite{britten1992analysis}), but not the human protocols. Deep RL agents also typically train quite slowly~\citep{lake2017building}. Furthermore, deep RL training algorithms are usually "model-free". That is, they resemble the slower neural processes underlying habit learning or perceptual learning, as opposed to the fast learning processes employed by humans to rapidly learn to perform these tasks in under an hour.

\section{Visual psychophysics of deep RL agents}

\subsection{Introduction}

To illustrate how Psychlab may be applied and its results interpreted, we offer a case study in which we apply the methods of visual psychophysics to probe visual behaviors of a state-of-the-art deep RL agent: UNREAL~\citep{jaderberg2016reinforcement}. Beyond its didactic value as an illustration for how Psychlab can be used, a study of basic visual behaviors of deep RL agents is of interest for several reasons. First, convolutional neural network architectures like the one in UNREAL have been motivated by arguments that they resemble the wiring of neural networks in visual cortex~\citep{fukushima1982neocognitron, riesenhuber1999hierarchical, lecun1995convolutional}. Second, convolutional neural network and primate performance on object recognition tasks has been studied extensively and compared to one another as well as to neural data recorded from several regions along the primate ventral stream~\citep{cadieu2014deep, khaligh2014deep, rajalingham2018large}. Third, several recent studies have also taken the perspective of probing convolutional neural networks along lines motivated by psychology and cognitive science but, unlike our study, mostly have used networks trained by supervised learning~\citep{zoran2015ordinal, ritter2017cognitive, NIPS2017_7146}. Such studies are limited by assumptions about how information can be read out and used in behavior. On the other hand, Psychlab works with complete agents. This makes it possible to study temporally extended visual behaviors. It also allows for a more direct comparison to human behavior on the exact same task. 

As we shall see later on, it turns out that UNREAL actually has \emph{worse} acuity than human, even when tested at the same spatial resolution. This result motivates a series of experiments on visually-guided pointing tasks leading to the surprising result that, unlike humans, this state-of-the-art deep RL agent is strongly affected by object size and contrast in ways that humans are not. It learns more quickly about larger objects. This then motivates us to introduce a model of foveal vision. Adding this mechanism to UNREAL turns out to improve performance on another family of labyrinth tasks that we didn't create ourselves: laser tag.

\subsection{What psychophysics has to offer AI}

\begin{figure}[h!]
\includegraphics[width=\textwidth]{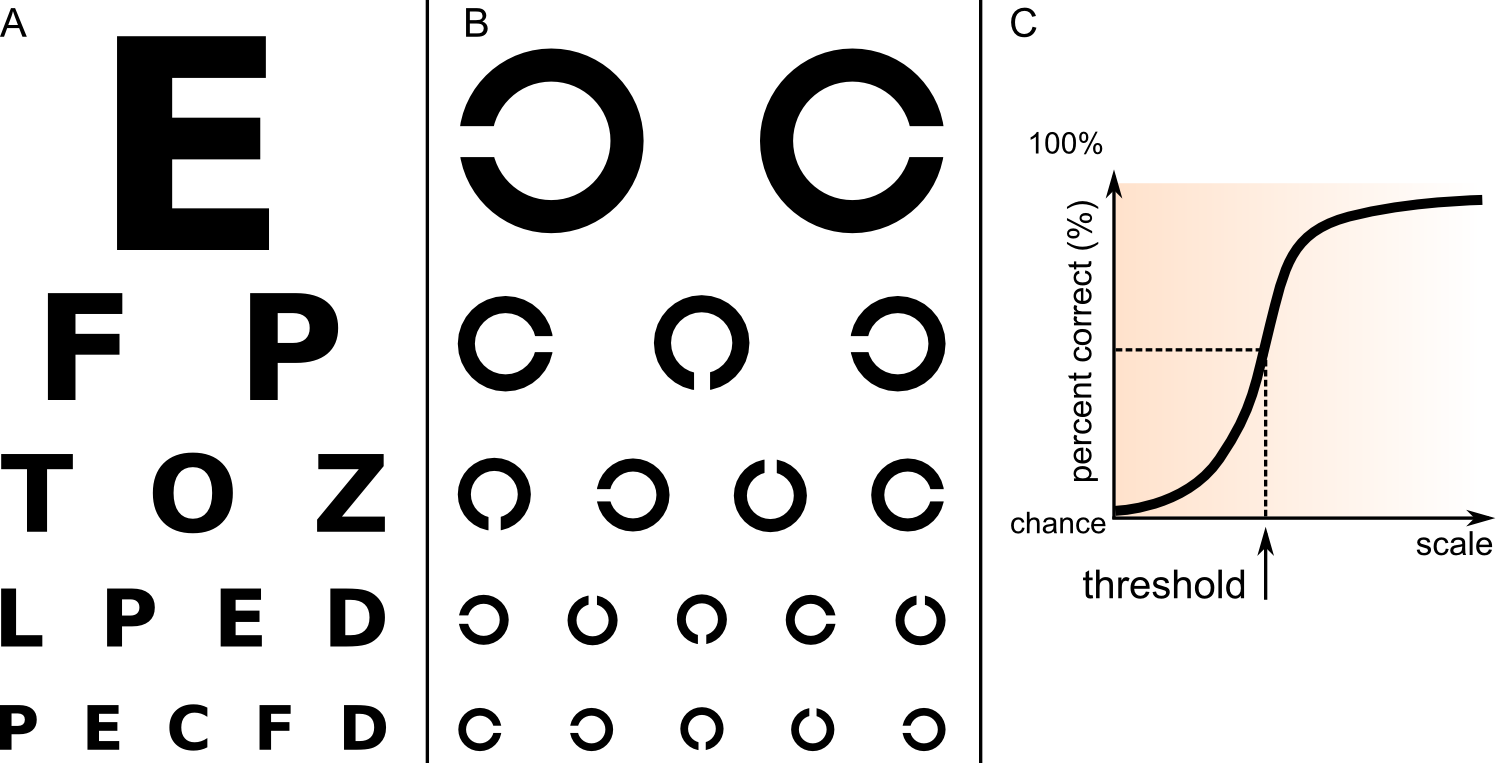}
\caption{Visual acuity measurement: (A) Snellen chart, acuity is determined by finding the smallest row of letters that can be correctly read aloud. (B) Landolt C chart, acuity is determined by finding the smallest row where the subject can correctly report the orientation of each optotype's opening. (C) Illustration of a psychometric curve for visual acuity measurement. The threshold scale is indicated}
\label{fig:eye_charts_and_threshold}
\end{figure}

Psychlab makes possible ways of analyzing experimental data that are common in psychology but relatively unknown in AI research. For example, we describe methods for measuring psychometric functions, detection thresholds, and reaction times for artificial agents that can be directly compared to those of humans.

The basic strategy of psychophysical measurement and analysis is to vary a critical stimulus parameter like luminance or contrast, and determine the threshold value required for the observer to perform a simple reference task \citep{farell1999psychophysical}. For example, one way to measure visual acuity is with a task that requires identification of a target stimulus that may be presented at a range of scales. For example, the target stimulus may be a Landolt C shape, the task then is to identify the orientation of the C's opening from a set of possibilities (see Fig. \ref{fig:eye_charts_and_threshold}-B). 

Measuring thresholds provides a principled way to compare task performances between agents or between agents and humans. A better way to report performance that will allow meaningful magnitude comparison is as follows. Thresholds are naturally in units of the stimulus or task parameters. Thus they are  immediately interpretable and actionable, unlike mean reward values. For example, one might have an application where an agent needs to be able to discriminate visual stimuli down to a specific small size $1/s$. Knowing that an agent's acuity is greater than $s$ immediately tells you that it is a candidate for that task while another agent with acuity $<s$ is not. This approach also makes sense for other tasks besides visual acuity. The only requirement is for the task to have meaningful difficulty parameters.

In addition, the shapes of psychometric functions for different task dimensions are informative. A flat psychometric curve tells you that a particular task/stimulus dimension has no impact on performance. Psychometric functions usually have a sigmoid shape~\citep{macmillan2004detection}.

\subsection{Methods}

\subsubsection{Adaptive staircase procedures}\label{section:staircase}
All the tasks we investigated included adaptive staircase procedures cf.~\citep{treutwein1995adaptive}. These worked as follows:

Let $r_1, \dots, r_K$ be a sequence of probability distributions over trials in increasing order of difficulty. For example, in a test of visual acuity, $r_0$ would be a distribution over trials where stimuli are displayed at the largest and easiest sized stimuli. $r_K$ would be a distribution over the smallest and most difficult sized stimuli. Let $r_c$, be the distribution of trials at difficulty level $c$. For each new trial, do one of the following (equiprobably): [base case] sample a trial from $r_c$, [advance case] sample a trial from $r_{c+1}$, or [probe case] first sample the difficulty level $r_p$ from $r_1, \dots, r_{\text{c}}$, then sample the specific trial from $r_p$. If after $c$ trials sampled $r_{c+1}$ (the advance case), the agent has achieved a score of 75\% correct or better then increment the base difficulty level to $c+1$. If the agent achieved a score worse than 50\% correct after $c$ trials sampled from $r_c$ then decrement the difficulty level to $c-1$.

Many tasks we investigated had two-dimensional adaptive staircase procedures. For example, the visual acuity and contrast sensitivity task simultaneously adjusted both scale and contrast parameters. These work by maintaining a separate difficulty level for each task dimension and equiprobably sampling advance type trials corresponding to independently incrementing one level or the other.

An episode---in the RL sense---consists of a fixed number of steps (3 minutes at 60 steps per second). Thus depending on its response time, an agent may complete a variable number of trials per episode. The adaptive staircase is reset to the initial difficulty level at the start of each episode.

For humans, this procedure feels like any other adaptive staircase. The task becomes easier or harder to adapt to the subject's performance level. For deep RL agents the exact same adaptive procedure induces a kind of curriculum. Early on in learning the agent does not complete many trials per episode, and when it does complete them it responds randomly. Thus it continues to experience the simplest version of the task. Only once it has learned to solve level $c$ then it gets level $c+1$. The probe trials ensure that trials at all easier difficulty levels remain interleaved. This simultaneously prevents catastrophic forgetting and smooths out the number of trials available for analysis at each difficulty level. 

\subsubsection{Human behavioral paradigms}

Human experiments were performed as follows. All experiments employed a within-subject design and  tested a single expert observer who was also an author of the paper. The Psychlab game window was used in full-screen mode on a $24^{''}$ widescreen monitor (HP Z24n 24-inch Narrow Bezel IPS Display). The subject was seated approximately 30cm away from the monitor. In the old version of the DM-Lab / Psychlab platform that was used for all experiments in this report, the size of the game window scaled automatically with the resolution. Thus for experiments at $640 \times 640$ resolution the game window was significantly larger than it was for experiments at $200 \times 200$ and $84 \times 84$. This limitation was removed in the new version of the platform that is being open sourced alongside this report. So it will be possible to vary resolution and size independently in future experiments with the open-source Psychlab platform.

\subsubsection{The \textit{UNREAL} agent}

The \textit{UNREAL} (UNsupervised REinforcement and Auxiliary Learning) agent~\citep{jaderberg2016reinforcement} combines the A3C (Asynchronous Advantage Actor-Critic)~\citep{mnih2016asynchronous} framework with self-supervised auxiliary control and reward prediction tasks. The base agent is a CNN-LSTM agent trained on-policy with the A3C loss. Observations, rewards and actions are stored in a small replay buffer. This experience data is used by the following self-supervised auxiliary learning tasks:
\begin{itemize}
    \item Pixel Control - auxiliary policies are trained to maximize change in pixel intensity of different regions of the input
    \item Reward Prediction - given three recent frames, the network must predict the reward that will be obtained in the next unobserved timestep
    \item Value Function Replay - further training of the value function is performed to promote faster value iteration
\end{itemize}

The \textit{UNREAL} algorithm optimizes a loss function with respect to the joint parameters, $\theta$, that combines the A3C loss $\mathcal{L}_{A3C}$ together with an auxiliary control loss $\mathcal{L}_{PC}$,
auxiliary reward prediction loss $\mathcal{L}_{RP}$ and replayed value loss $\mathcal{L}_{VR}$, with loss weightings $\lambda_{PC}$, $\lambda_{RP}$, $\lambda_{VR}$:

$\mathcal{L}_{UNREAL}(\theta) = \mathcal{L}_{A3C} + \lambda_{PC} \mathcal{L}_{PC} + \lambda_{RP} \mathcal{L}_{RP} + \lambda_{VR} \mathcal{L}_{VR}$

We note that the true objective function that we would like to maximize is the (undiscounted) score on the tasks, and the loss function is merely a proxy for this. In particular, we consider the discount parameter $\gamma$ to be an agent-side hyperparameter, not a parameter of the environment.

\subsection{Experiments}

Note: All experiments in this section were performed with an earlier version of the Psychlab and DM-Lab codebase than the version in the open-source release. The differences between the old platform and the new one are minor and unlikely to influence the results.

\subsubsection{Visual acuity and contrast sensitivity}

A video of human performance on the test of visual acuity and contrast sensitivity can be viewed at \url{https://youtu.be/yWA1-hFU9F0} (Landolt C discrimination).

Unlike resolution, which simply denotes the number of pixels in an image, visual acuity refers to an agent's ability to detect or discriminate small stimuli. Greater acuity means greater ability to perform a reference task despite scaling the stimuli down to smaller and smaller sizes. Input image resolution and visual acuity may not be related to one another in a simple way.

In primates, visual acuity and contrast sensitivity are immature at birth. Acuity develops over the course of the first year of life, primarily due to peripheral mechanisms including the migration of cones to the fovea \citep{wilson1988development, kiorpes1998peripheral}.

\begin{figure}[h!]
\includegraphics[width=\textwidth]{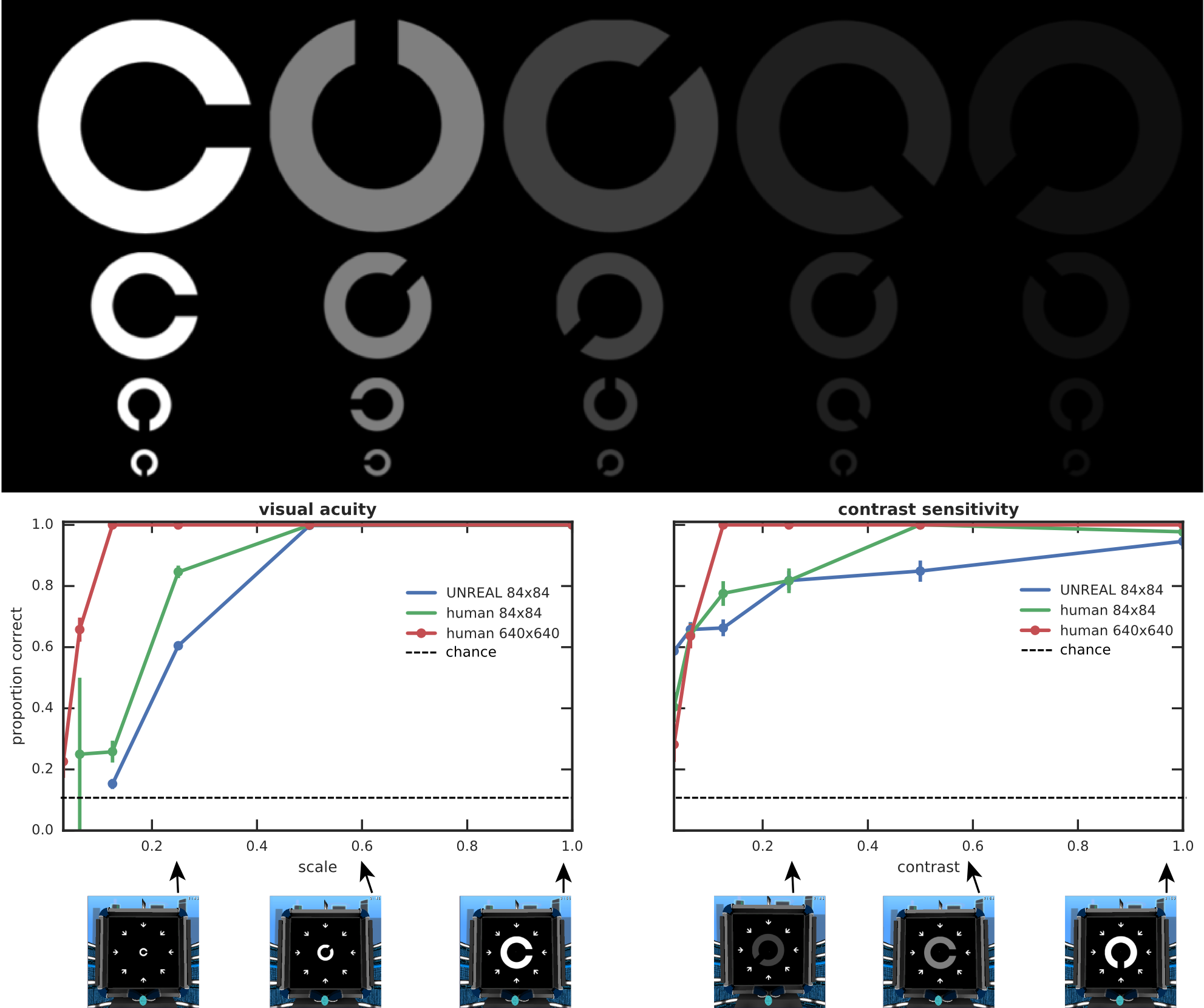}
\caption{Comparison of UNREAL and human visual acuity. Here  visual acuity was measured by Landolt C orientation discrimination accuracy. The scale at which the Landolt C stimulus was displayed was adapted according to the staircase procedure described in Section \ref{section:staircase}. As can be seen, even when the input \emph{resolution} is the same for humans and agents ($84\times 84$) visual \emph{acuity} performance can differ.}
\label{fig:acuity}
\end{figure}

To test the visual acuity of the UNREAL agent, we trained it to identify Landolt C stimuli presented at a range of scales. The resulting psychometric function (accuracy as a function of stimulus scale) had a sigmoid shape that qualitatively matched the shape obtained when measured for a human observer (Fig. \ref{fig:acuity}). We also measured the UNREAL agent's contrast sensitivity with the Landolt C identification task by presenting the stimuli at a range of contrasts. In this case the psychometric function (accuracy as a function of stimulus contrast) differed qualitatively from human. UNREAL was less accurate than human for moderate contrasts, but it outperformed human level at the lower contrast levels (Fig. \ref{fig:acuity}). 

Human observers performed the visual acuity and contrast sensitivity tasks with Psychlab running at two different resolutions: $640 \times 640$ (high resolution) and $84 \times 84$ (low resolution). Since the standard implementation of UNREAL runs at $84 \times 84$, i.e. it downsamples to that size, it is not surprising that its acuity was significantly worse than the result obtained from a human observer at $640 \times 640$. However, the more surprising finding was that UNREAL also had worse acuity relative to the human observer at $84 \times 84$ (Fig. \ref{fig:acuity}). A similar result was obtained with contrast sensitivity. Human contrast sensitivity at $640 \times 640$ was greater than at $84 \times 84$, but UNREAL performed worse than both at moderate contrast levels (Fig. \ref{fig:acuity}).

Why did UNREAL have worse visual acuity than the human observer, even when both viewed the stimuli at $84 \times 84$? The fact that humans perform better means that there must be additional target-related information available in the observation stream, but UNREAL cannot take advantage of it. One possibility is that this information is aliased by the subsampling that happens between convolutional layers in the standard UNREAL network  architecture.

One surprising conclusion to draw from these experiments is that  effective acuity cannot be inferred from image resolution alone.

\subsubsection{Global form perception: Glass patterns}

A video of human performance on the Psychlab implementation of the Glass pattern detection task can be viewed at \url{https://youtu.be/XLQ9qgV9qrE}.

\begin{figure}[h!]
\includegraphics[width=\textwidth]{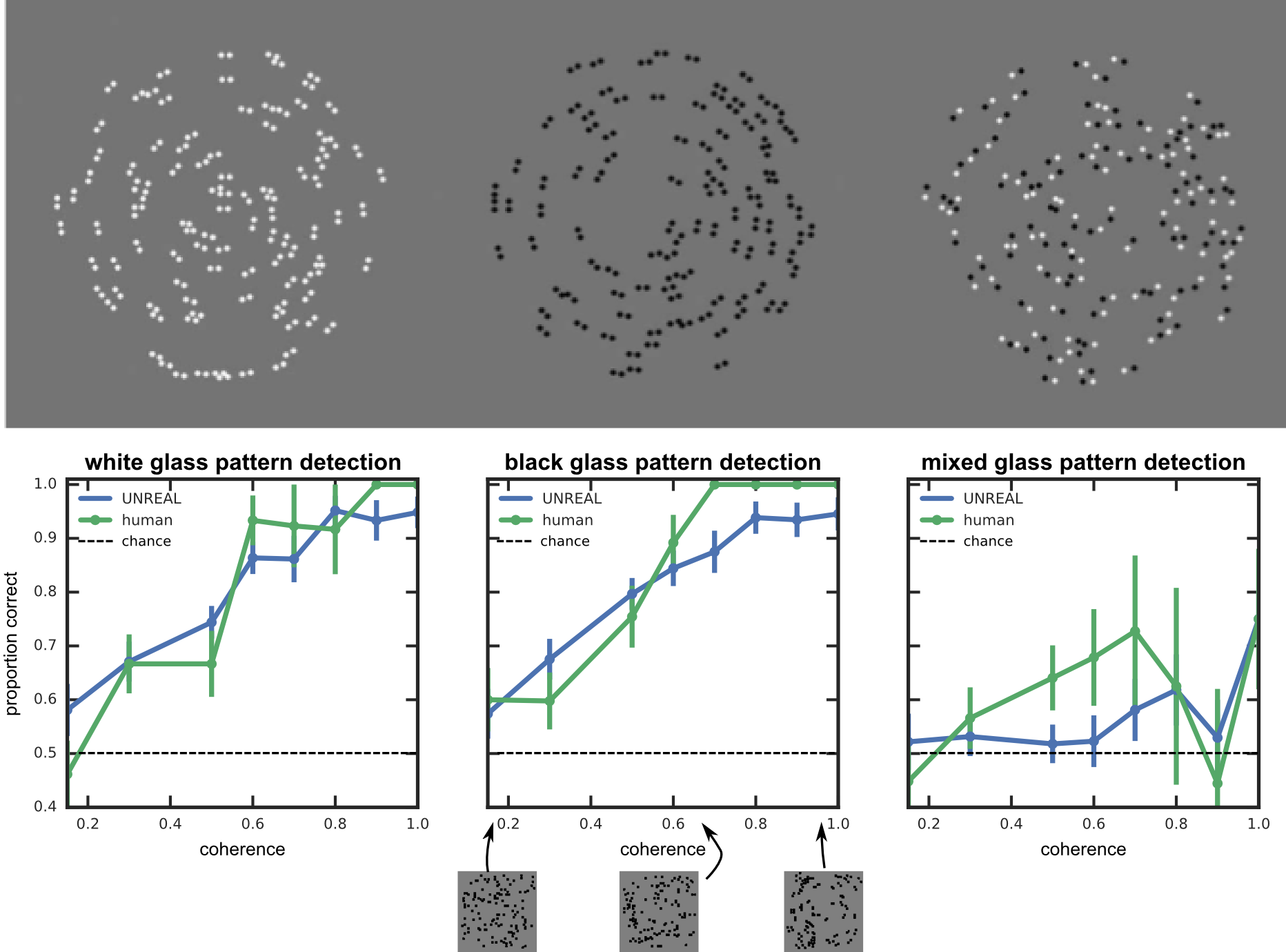}
\caption{Glass pattern example stimuli (modified from~\cite{badcock2005interactions}) and psychometric curves for UNREAL $(84 \times 84)$ and our human observer (also at $84 \times 84$).} See text for details of the experimental setting. As can be seen, UNREAL and humans perform very similarly under the different experimental settings.
\label{fig:glass_patterns}
\end{figure}

In the previous experiment we measured how well UNREAL coped with small visual stimuli and found it to have impaired acuity relative to human, even when tested at the same resolution. Next we set out to determine whether this insensitivity to small stimulus features may have implications extending to larger object perception. Specifically, we studied perception of Glass patterns, a class of stimuli where a strong impression of global form arises from high spatial frequency cues (Fig. \ref{fig:glass_patterns}).

A Glass pattern stimulus is a random dot pattern where each dot is paired with a partner at a specific offset~\citep{glass1969moire, glass1973perception}. The pair together are called a dipole. In a concentric Glass pattern, dipoles are oriented as tangents to concentric circles. Glass patterns are built from local correspondences yet they evoke a percept of global structure. Their perception must involve at least two stages. First local features must be extracted (dipole orientations). The resulting orientation information coming from different parts of the image must be integrated in order to represent global form. Glass patterns have been used in numerous studies of human form perception e.g.~\citep{glass1976pattern, dakin1997detection, wilson1998detection} and used in neuroimaging~\citep{mannion2009discrimination} as well as single-unit electrophysiology~\citep{smith2002signals, smith2007glass} studies. From the perspective of deep RL, Glass pattern stimuli are particularly nice stimuli because an infinite number of different ones can easily be produced. This property is useful for excluding the possibility that an agent might overfit to a specific set of stimuli.

Global form perception as measured by Glass pattern detection is impaired in a number of disorders including autism~\citep{simmons2009vision} and migraine~\citep{mckendrick2006vernier}.

We measured human and UNREAL performance on a Glass pattern detection 2AFC task~\citep{macmillan2004detection}. In each trial, one Glass pattern stimulus (the target) was paired with a distractor pattern created by placing dots randomly, preserving average inter-dot distances. We measured a psychometric function by progressively degrading the pattern by injecting additional noise dots, thereby progressively lowering pattern coherence.

Glass pattern detection psychometric function were similar for UNREAL and our human observer. Both UNREAL and the human observer had sigmoid shaped psychometric functions for the detection of both white and black Glass patterns on a gray background. In both cases, the threshold, defined as the coherence level where detection performance rose above $75\%$, was around $0.5$ (Fig. \ref{fig:glass_patterns}). 

Interestingly, both UNREAL and the human observer were unable to reliably detect mixed polarity Glass patterns (Fig. \ref{fig:glass_patterns}). These Glass patterns are created from dipoles consisting of one white and one black dot. This effect is known for human observers~\citep{wilson2004glass, barlow2004convergent, badcock2005interactions, burr2006effects}. However, this is likely the first time it has been described in an artificial system not created for this specific purpose. 

How is it possible that UNREAL has human-like Glass pattern detection thresholds despite having worse than human acuity? It is common to conceptualize Glass pattern perception as depending on a two-stage process. First local orientation information must be extracted from each dipole then, second, orientation information coming from different parts of the image is integrated to represent global form. UNREAL's weak visual acuity may be expected specifically to impact the first of these stages. How then can it recover normal human-level performance in the second stage if information has already been lost in the first stage? One possibility is that the acuity loss only makes dipole detection less reliable, but when integrating over the entire stimulus it is still possible to obtain enough orientation information that the global percept can be represented despite accumulating errors from individually misperceived dipole orientations.

\subsubsection{Point to target task}\label{section:point_to_target}

One particularly unbiological aspect of most modern artificial neural networks, including UNREAL, is weight sharing. For reasons having to do with computational efficiency,  convolutional neural networks sum incoming gradients over the entire visual field, and use the resulting signal to adjust the weights of all cells, no matter what part of the visual field they are responsible for\footnote{In contrast, biological neurons are presumably tuned using signals that are more local to their receptive field. This may explain heterogenous spatial sensitivity patterns like those described by~\cite{afraz2010spatial}. Weight sharing convolutional networks would not be able to explain such data.}. In this section we ask whether the convolutional weight sharing structure leads to abnormalities in visual learning.

\begin{figure}[h!]
\includegraphics[width=\textwidth]{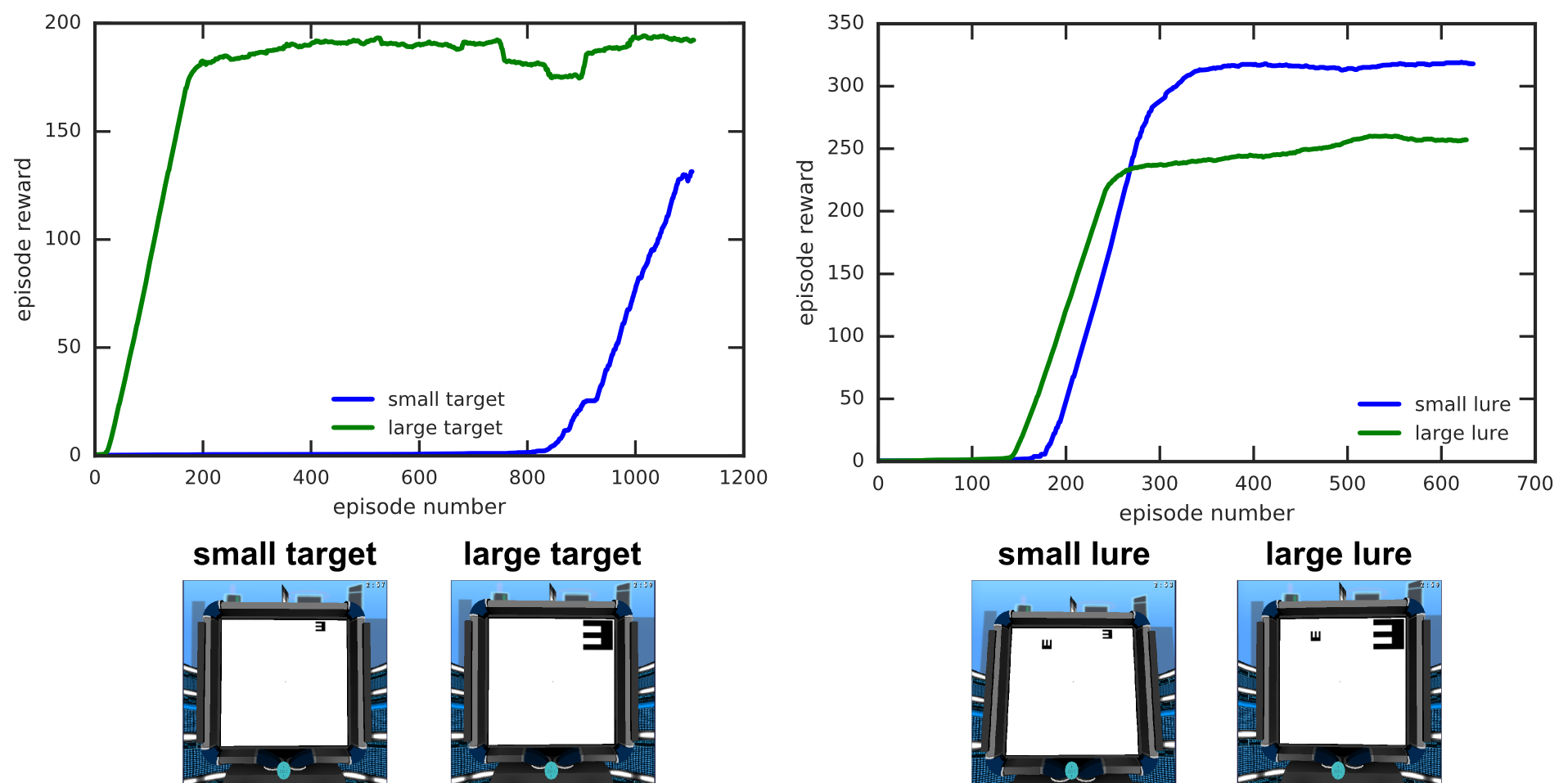}
\caption{Point to target task - the agent needs to find a target which is either small or large on the screen, without (left) or with (right) the presence of a lure (small or large). UNREAL seems to be quite sensitive to the size of the target and lure. As can be seen, learning is much faster when the target is large. Additionally, the presence of a large lure hurts final performance significantly.}
\label{fig:point-to-target}
\end{figure}

We trained UNREAL on a pointing task. Each trial was initiated by foveating a fixation target. Upon initiation, a target stimulus (the letter 'E') appears at some distance away, in one of two possible locations. After pointing to the target, the trial ends and a reward is delivered. There were two versions of the task that differed only in the size of the target. 

UNREAL learns the large-target version of the task more quickly than it learns the small-target version (Fig. \ref{fig:point-to-target}). That is, it requires less interactions with the environment to achieve perfect performance. This may be due to weight sharing. In a convolutional neural network the gradient signal is summed over space, thus, when a larger region of the visual field is correlated with the loss signal, then gradient descent steps will be larger. In the case of online reinforcement learning algorithms like UNREAL, gradient descent steps are time-locked to environment steps, so this implies that the larger the region of the visual field that correlates with the reward, the bigger the summed gradient becomes, and thus the faster learning can proceed.  This interpretation implies that the effect is not peculiar to UNREAL. Any deep RL agent that time-locks network updates to environment steps and employs a convolutional neural network to process inputs will show the same effect.

Next, we studied two more variants of the pointing task in which two objects appeared in each trial. In both cases pointing to the object on the left (the target) yielded a reward of 2 while pointing to the target on the right (the lure) provided a reward of 1. A trial ends when the agent points to either the target or the lure. In one version of the task both target and lure were the same size. In the other version the lure was larger. 

Since UNREAL learns more quickly about larger objects, we hypothesized that the presence of the larger lure stimulus would make it likely to become stuck in the suboptimal local optimum corresponding to a policy that exclusively points to the lure rather than the target. This is indeed what occurs. It tends to find the policy of looking to the target when both target and lure are the same size but tends to find a policy of pointing to the lure when the lure is larger (Fig. \ref{fig:point-to-target}). Psychlab made it easy to detect these subtle distortions of the learning process arising from weight sharing in convolutional neural networks, but the underlying issues are unlikely to be limited to Psychlab. It is likely that UNREAL frequently gets stuck in suboptimal local optima as a result of its learning first about larger objects then smaller objects. In more complicated tasks this could play out in countless subtle ways.

\subsubsection{Foveal vision and the effects of input resolution}

The input resolution of a deep RL agent is a rather critical hyperparameter. For example, it affects the number of units in each level of the network, which in turn affects the speed that the whole system can be operated and trained. Optimal sizes of convolutional filters also depend strongly on the input resolution. Thus it is not always a simple matter to change it without also needing to simultaneously change many other aspects of network design. 

One could argue it was unfair to limit the human subject to $84 \times 84$ since they could have performed the task at higher resolutions and thereby achieved higher scores, at least on the tasks that require resolving small objects. $84 \times 84$ was only chosen because it was the resolution UNREAL was designed for. Moreover, you can't easily just run UNREAL at a higher resolution. At minimum you'd have to change the filter sizes in the convolutional network. This could easily either make the algorithm too slow to run or break it altogether. 

\begin{figure}[h!]
\includegraphics[width=\textwidth]{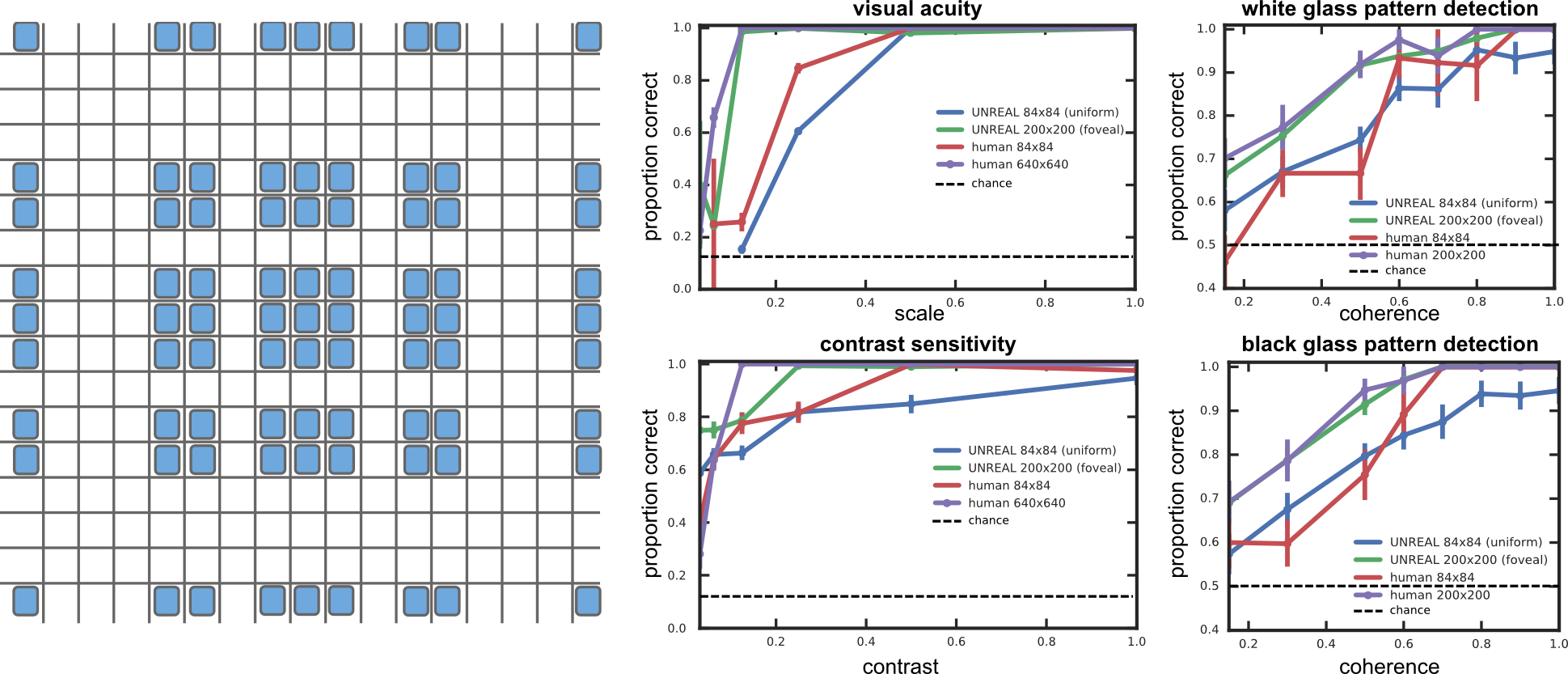}
\caption{(A) Fovea illustration. Cells are pixels in the original image, blue cells are pixels kept after processing with the fovea model.  Other subfigures will be psychometric curves comparing fovea to no-fovea and human (both $84 \times 84$ and $200 \times 200$) for (B) acuity, (C) contrast sensitivity, and (D) glass patterns.}
\end{figure}

Thus we introduced a simple "fovea" model in order to scale up UNREAL to higher resolutions without changing its neural network architecture or making it run more slowly.

To emulate a fovea, we subsample the image by keeping only a fixed subset of all the rows/columns of pixels composing the image, and discarding the others. We choose this subset so that for a given image, as we move away from the centre along the vertical (resp. horizontal) axis, more and more rows (resp. columns) are discarded. For example, consider an original image of size $101 \times 101$, with pixel values $O(i,j)$ for $i,j \in [-50,+50]$ with $i,j \in \mathbb{Z}$.\footnote{We use $[-50,+50]$ so the origin is at (0,0).} Say we want to subsample down to $11\times 11$ according to the fovea model. The final image pixel values will be $F(u,v) = O(\sigma(u), \sigma(v))$ for $u,v \in [-5, +5]$, where we choose $\sigma:\mathbb{Z}\to\mathbb{Z}$ to be a strictly increasing, anti-symmetric function such that: 
$$\begin{cases}
i < j \implies \sigma(i) < \sigma(j) \\
\sigma(-i) = -\sigma(i) \\
\sigma(5) \leq 50 \\
\sigma \text{ grows polynomially: } \sigma(\lvert x \rvert) \sim x^8 \\
\end{cases}$$

This model is of course only a crude approximation to the fovea of the human eye. First, we don't have a distinction between "rods" and "cones" which have vastly different distributions across the retina and different response characteristics \citep{purves2001neuroscience}. Second, the sampling density in the retina changes quite rapidly as we move further from the fovea, while our sampling density decays somewhat slower. Nevertheless we argue that qualitatively this model captures the most salient property of the fovea, sampling the central parts of the visual field with higher density thereby increasing the acuity in those parts. 

\subsubsection{Foveal vision for standard DM-Lab laser tag tasks}

\begin{figure}[h!]
\includegraphics[width=\textwidth]{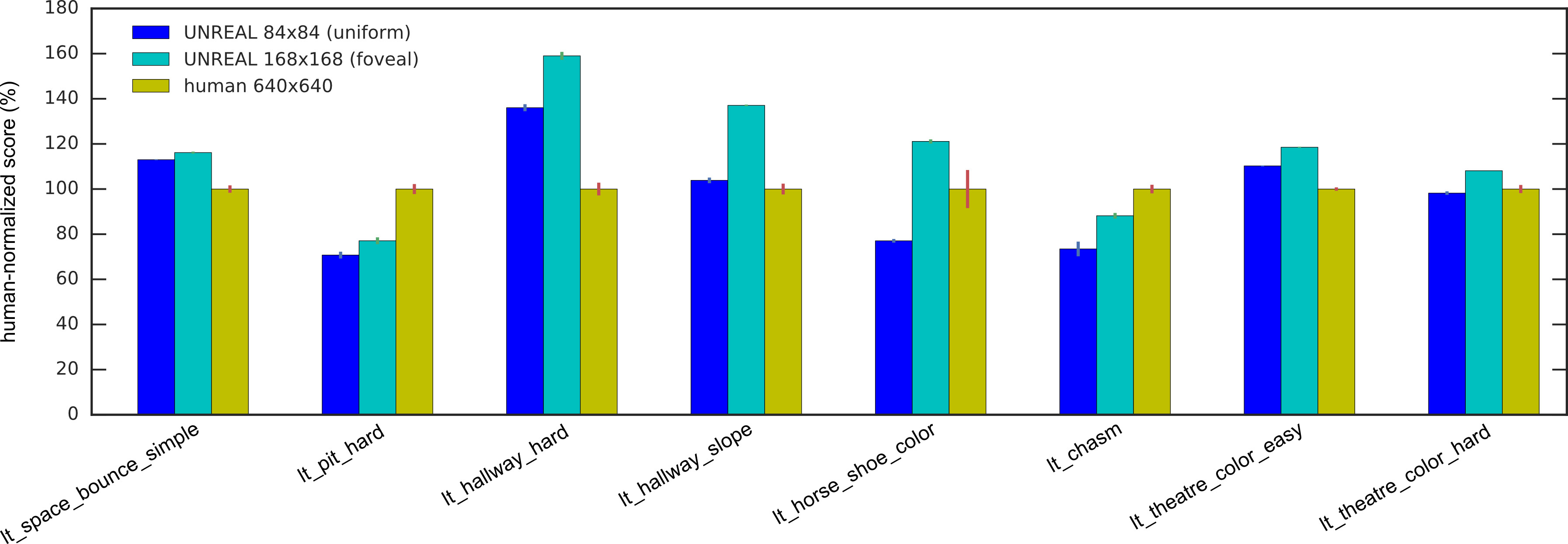}
\caption{Laser tag: bar graph comparing UNREAL performance with and without fovea to human level. The foveated UNREAL agent outperforms the unfoveated agent in all levels.}
\label{fig:laser_tag}
\end{figure}

This fovea model has the effect of expanding the input representation of objects located in the center of the visual field. Now recall our result from the point to target task (see section~\ref{section:point_to_target}), UNREAL learns quicker when larger objects---in terms of their size on the input image---are correlated with reward. This suggests that the fovea model should improve learning speed or overall performance on tasks where the agent typically centers its gaze on the objects that are most correlated with reward. Fortunately, quite a few DM-lab tasks already exist that have this property: all the laser tag tasks. We compared the performance of UNREAL running at its default resolution of $84 \times 84$ to UNREAL with inputs rendered at $168 \times 168$ and foveally downsampled to $84 \times 84$. The fovea model turned out to improve performance on all eight laser tag tasks (Fig.~\ref{fig:laser_tag}).

\subsubsection{Visual search}

Visual search may ultimately prove to be the best use-case for Psychlab for several reasons.
\begin{enumerate}
    \item Human visual search behavior is extremely well-characterized. Thousands of papers have been written on the topic (for a recent review see \cite{wolfe2017five}). Considerable detail is also known about its neural underpinnings~\citep{desimone1995neural, maunsell2006feature, bisley2010attention}. Such detail with which phenomena related to visual search behavior are understood provides the ideal setting for productive Psychlab experiments. Psychlab makes it easy to test whether deep RL agents reproduce the same behavior patterns. The original human experiments can be replicated through Psychlab so that results from humans and deep RL agents can be displayed side by side. 
    
    \item It is a true visual \emph{behavior}. It unfolds over time. In fact, reaction time is typically the dependent measure used in human experiments. Since time is so central, visual search behavior is not a natural fit for discriminative modeling. While progress has been made in addressing its "bottom-up" stimulus-driven aspect and predicting the first few fixations~\citep{itti2000saliency, bruce2015computational}, these models do not do a good job predicting task-driven search behavior~\citep{wolfe2017five}. We suspect that an accurate model will require a full RL-based agent integrating stimulus-driven information with top-down feedback based on the recent search history and task goals to actually search the display.
    
    \item A major question in this area, "serial" vs. "parallel" processing, appears to reflect a suboptimality in human perceptual intelligence. We know that convolutional networks can be constructed that are capable of localizing arbitrary features in constant time, assuming parallel computational of filter responses as on a GPU. Why then did evolution not find such a solution? This is especially puzzling since some visual searches do appear to operate in such a parallel manner, they are said to "pop-out", while other searches take longer and longer the more items there are to be searched through~\citep{treisman1980feature}. Moreover, the subjective phenomenon of serial processing appears suspiciously close to the very essence of thought. Is it really a suboptimality? Or do the apparently serial  visual search effects hint at something fundamental about neural computational processes that we have yet to understand?
    
    \item Studying visual search behavior is one of the main accepted ways to approach the cognitive faculty we call attention. In supervised deep learning, considerations of "attention" or other methods of restricting the size of hypothesis spaces have proven essential for scalability toward realistically sized images and language corpuses~\citep{bahdanau2014neural, ba2014multiple, xu2015show, gregor2015draw}. Deep RL has not yet developed to the point where such scaling up has been seen as important, but it is likely to encounter the same issues. Remember, the current state-of-the-art systems operate at $84 \times 84$ and cannot easily be scaled up\footnote{There has been recent work on using RL to guide the deployment of an attention model in order to improve supervised and unsupervised learning~\citep{ba2014multiple, xu2015show}, but there has not yet been much work that models attention specifically to speed up deep RL itself (cf.~\cite{mnih2014recurrent}): attention for RL as opposed to RL for attention.}.
\end{enumerate}

\begin{figure}[h!]
\includegraphics[width=\textwidth]{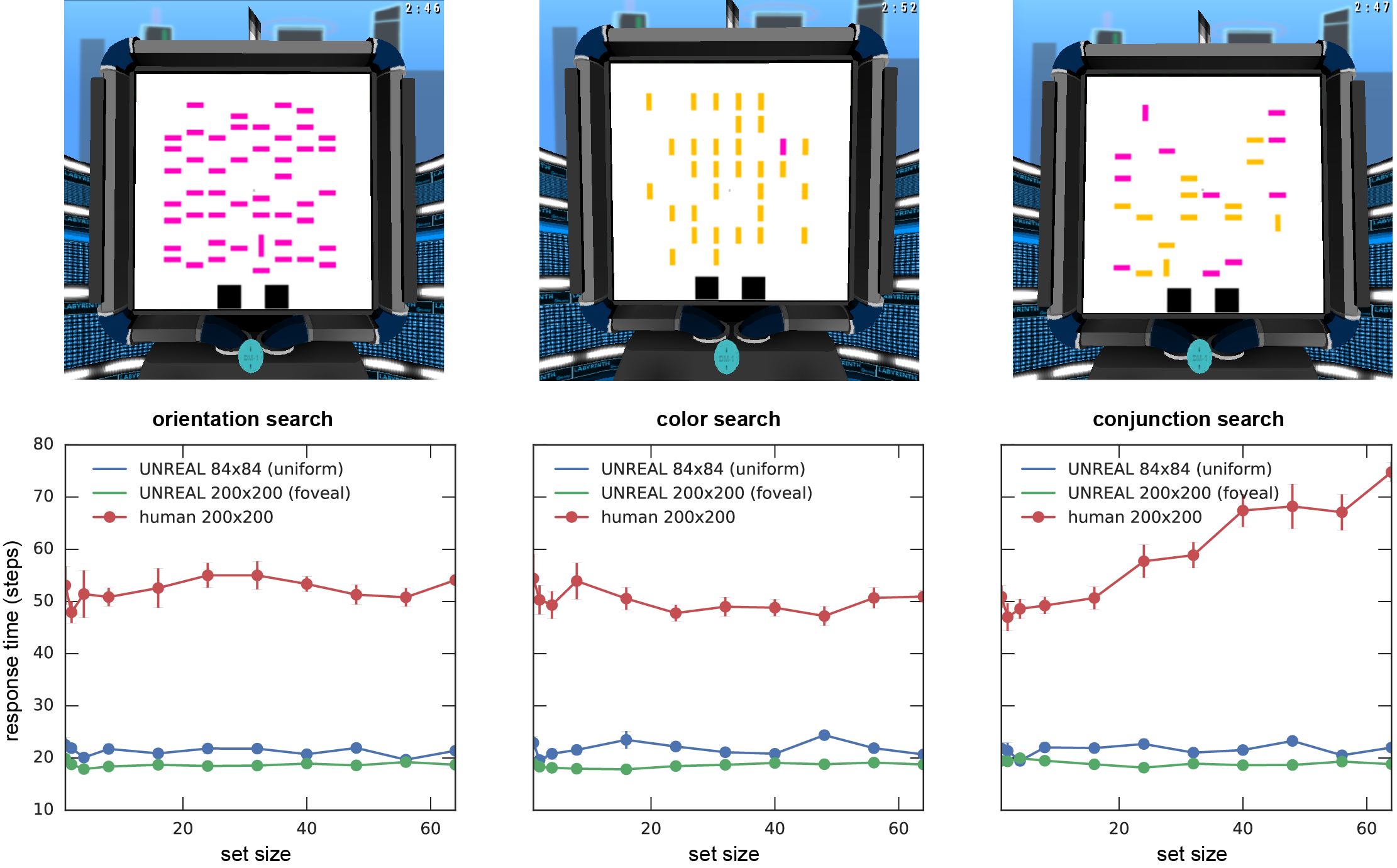}
\caption{Visual search reaction times: orientation search, color search, and conjunction search. Human reaction times for orientation search and color search are independent of set size but scale linearly with set size for conjunction search. This replicates a well-known result in the literature~\citep{treisman1980feature, wolfe2017five} and validates the Psychlab implementation of this task. Unlike the human pattern of results, UNREAL's reaction time is  independent of set size in all three cases.}
\label{fig:visual_search}
\end{figure}

Replicating the classic human result, we show that human reaction times for orientation search and color search are independent of set size while reaction times for conjunction search scale linearly with set size (Fig.~\ref{fig:visual_search}). This validates our Psychlab implementation of visual search paradigm by showing it produces a pattern of human results that replicates that achieved with other task implementations.

UNREAL is also able to solve this task. It performs almost perfectly after training. Human performance is also near perfect. However, that is where its resemblance to human visual search behavior ends. As expected for an architecture that processes visual inputs with a convolutional neural network, UNREAL's reaction time is always independent of set size, even for conjunction search (Fig.~\ref{fig:visual_search}).

\subsubsection{Motion discrimination}

A video of human motion discrimination task performance in Psychlab can be viewed at: \url{https://youtu.be/IZtDkryWedY}.

Random dot motion discrimination tasks are used to investigate motion perception in the presence of noise~\citep{newsome1990neuronal, britten1992analysis}. Subjects must discriminate the direction in which a majority of dots are moving. The fraction of dots moving coherently versus randomly can be varied by the experimenter in order to determine motion perceptual thresholds. Lesions of cortical area MT impair performance, increasing motion discrimination thresholds measured by this task~\citep{newsome1988selective}. Going beyond the specific study of motion perception, this task has also been important in the study of perceptual decision-making more generally~\citep{britten1996relationship, shadlen2001neural}.

In primate data, there is typically a threshold at which the subject's ability to discriminate motion direction declines sharply. Reaction time generally increases near the perceptual threshold. Systematic relationships between reaction time and stimulus parameters have been taken as evidence for diffusion-to-bound models of perceptual decision-making~\citep{ratcliff1998modeling, hanks2006microstimulation, shadlen2006speed, hanks2014neural}.

We found that UNREAL failed to learn to perform this task at any level of motion coherence (Fig.~\ref{fig:additional_tasks}). It's possible that additional steps corresponding to the operant conditioning steps used in the animal training paradigms should be added to the protocol. For example, when training macaques to perform the motion discrimination task, \cite{britten1992analysis} employed a curriculum of increasingly difficult to learn subtasks: (1) fixation only (2) saccade to a single target (3) saccade to a single target in the presence of random dot motion patterns, and (4) choose the saccade target based on the direction of coherent motion. A similar curriculum may work with deep RL agents. 

Note: the version of the motion discrimination task used in the results reported here differed slightly from the version included in the Psychlab open-source release. Instead of flashing random incoherent dots at different locations as described in the protocol of \cite{britten1992analysis}, the incoherent dots moved in random directions different from the overall direction of motion for the experiments reported here. This difference is unlikely to qualitatively change the results~\citep{pilly2009difference}.

\subsubsection{Change detection}
A video of human performance on the Psychlab implementation of the change detection task can be viewed at \url{https://youtu.be/JApfKhlrnxk}.

In the change detection task \citep{phillips1974distinction, luck1997capacity} the subject viewed a sample array of objects and a test array on each trial, separated by a delay (the retention period). The task was to indicate whether the two arrays were identical or whether they differed in terms of a single feature. The objects we used were squares or letter-E shapes (equiprobably) at different colors and orientations. This task used a two-interval forced choice design~\citep{macmillan2004detection}, i.e. a sequential comparison design. Task performance in humans drops as the number of objects is increased~\citep{luck1997capacity}. Performance is also dependent on the retention period---longer delays between the two sets hurt performance more. This task is regarded as measuring the capacity or fidelity of visual working memory. It has been influential in the debate over whether human visual working memory has a discrete item limit or whether it is more flexibly divisible~\citep{alvarez2004capacity, awh2007visual, zhang2008discrete}. Interestingly, in humans, individual differences in change detection tests of visual working memory are highly correlated with fluid intelligence~\citep{fukuda2010quantity, luck2013visual}.

UNREAL fails to perform the change detection task, regardless of delay period or set size (Fig.~\ref{fig:additional_tasks}). This tells us that despite its having an LSTM, its ability to learn to use it in a manner akin to human visual working memory is quite limited.

\subsubsection{Multiple object tracking}

A video of human performance on the Psychlab implementation of the multiple object tracking task can be viewed at \url{https://youtu.be/w3ddURoeQNU}.

In the multiple object tracking task~\citep{pylyshyn1988tracking, cavanagh2005tracking}, subjects are presented with a display on which several indistinguishable objects (here circles) are moving. At the beginning of a trial, some of the objects are briefly identified as the target set (by a temporary color change). At the end of the trial, one of the objects is identified as the query object (by color change). The subject's task is to report whether or not the query object was from the target set. The experimenter can vary the number of objects, the number in the target set, and the speed of movement.

The task probes the subject's ability to attend to multiple objects simultaneously. In human trials, most subjects are able to track four or five objects. This also suggests---since the objects are indistinguishable---that there is an indexical aspect to human attention~\citep{pylyshyn2001visual}.

\begin{figure}[h!]
\includegraphics[width=\textwidth]{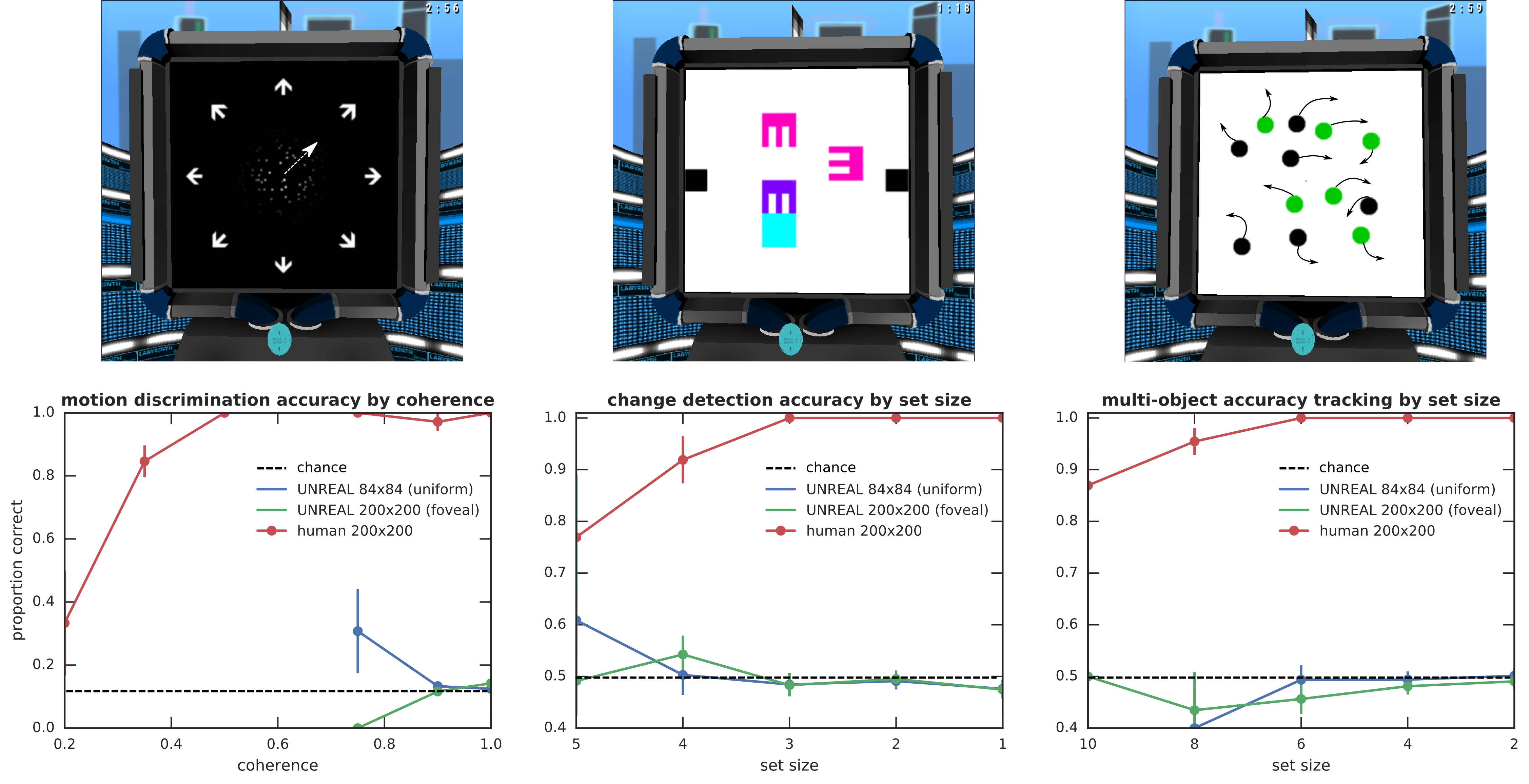}
\caption{Psychometric functions comparing human and UNREAL on motion discrimination, change detection, and multiple object tracking tasks. UNREAL is not able to perform these tasks, performing at chance levels under all settings.}
\label{fig:additional_tasks}
\end{figure}

UNREAL also fails to perform the multiple object tracking task (Fig.~\ref{fig:additional_tasks}). Taken alongside the results on motion discrimination and change detection tasks, this provides further evidence that, despite its LSTM, UNREAL has considerably difficulty learning to integrate information over time.

\section{General discussion}

This report introduces Psychlab, a DM-Lab-based platform enabling psychology-style behavioral experimentation with deep RL agents. Psychlab is intended to complement other environments for Deep RL research like the standard DM-Lab levels~\citep{beattie2016deepmind} and VizDoom~\citep{kempka2016vizdoom}. Psychlab maintains the same agent interface as DM-Lab while making it much easier to perform experiments that emulate behavioral paradigms from cognitive psychology or psychophysics. The Psychlab task definition API is easy-to-use and highly flexible, as demonstrated by the range of paradigms we are making available along with this report\footnote{Available on github:  \url{insert_url_here}. The current list is: visual acuity and contrast sensitivity measurement, glass pattern detection, visual search, sequential comparison (change detection), random dot motion discrimination, arbitrary visuomotor mapping, continuous recognition, and  multi-object tracking.}. Psychlab also makes it easy to directly compare humans and artificial agents. Human experiments with Psychlab replicate classical effects from the literature like the efficiency of orientation search and color search---reaction time independent of set size---versus the inefficiency---linearly increasing reaction time with set size---for the corresponding conjunction search~\citep{treisman1980feature, wolfe2017five}.

As an example for how Psychlab can be applied to deep RL, we undertook a study of the visual psychophysics of the UNREAL deep RL agent~\citep{jaderberg2016reinforcement}, in comparison to human. We found that UNREAL was affected by target size and contrast in a similar manner to human observers, similarly shaped psychometric functions. However, we found that UNREAL's visual acuity was worse than human-level, even when the human comparison was limited to observing Psychlab at the same, quite low, input resolution as UNREAL $(84 \times 84)$. On the other hand, UNREAL's global form perception, as measured by a Glass pattern detection task, was more similar to the human pattern of results. In particular, unipolar Glass pattern detection had a similarly shaped psychometric function for human observers and UNREAL while mixed polarity Glass patterns were found to be difficult to detect for both. On a visual search task, UNREAL did not show the classical reaction time effects that characterize human experiments with this paradigm.

These results on the visual psychophysics of UNREAL motivated us to test whether UNREAL might learn more quickly to recognize larger target objects. This turned out to be the case. We also showed that this effect can sometimes cause learning to become stuck in suboptimal local optima corresponding to larger objects despite the presence of smaller more rewarding objects. Its likely that these effects play out in numerous subtle ways in other deep RL tasks. Psychlab just makes it easy to measure them. Another implication of this result was that an input preprocessing step that expands the center of the observation---inspired by foveal vision---should improve performance on tasks where the important objects tend to be at the center of gaze. We tested this hypothesis with standard DM-lab laser tag tasks and found that adding this "foveal" preprocessor to UNREAL improved performance on all eight available laser tag levels. This story demonstrates how probing the details of agent behavior with Psychlab can feed directly back into deep RL research to yield new ideas for improving agent performance.

\section*{Acknowledgements}
First, we would like to thank Gadi Geiger and Najib Majaj for kindling our interest in psychophysics. We also thank Chris Summerfield, Dharshan Kumaran, Koray Kavukcuoglu, Thore Graepel, Greg Wayne, Vlad Mnih, Sasha Vezhnevets, Simon Osindero, Karen Simonyan, Ali Eslami, Carl Doersh, Jane Wang, and Jack Rae for the many insightful and influential conversations we had while conceptualizing this work. Additionally, we would like to thank Stig Petersen, Tom Handley, Vicky Langston, Helen King, and Adrian Bolton for project management support, Marcus Wainwright for art, and Aliya Ahmad for comms.

\bibliographystyle{plainnat}
\bibliography{biblio}

\end{document}